\documentclass[10pt,twocolumn,letterpaper]{article}

\usepackage{cvpr}  
\definecolor{cvprblue}{rgb}{0.21,0.49,0.74}
\usepackage[pagebackref,breaklinks,colorlinks,allcolors=cvprblue]{hyperref}
\usepackage{amsmath}
\usepackage{algorithm}
\usepackage{algpseudocode}
\usepackage{multirow}
\usepackage[normalem]{ulem}

\title{Neural Architecture Search of Time-to-First-Spike-Coded Spiking Neural Networks for Efficient Eye-based Emotion Recognition}

\author{
    Qianhui Liu$^{1}$ \quad
    Jing Yang$^{2}$ \quad
    Miao Yu$^{3}$ \quad
    Trevor E. Carlson$^{3}$ \quad
    Gang Pan$^{4}$ \\
    Haizhou Li$^{5}$ \quad
    Zhumin Chen$^{1}$ \\
    $^{1}$School of Artificial Intelligence, Shandong University, Jinan, China \\
    $^{2}$School of Software, Shandong University, Jinan, China \\
    $^{3}$School of Computing, National University of Singapore, Singapore \\
    $^{4}$College of Computer Science and Technology, Zhejiang University, Hangzhou, China \\
    $^{5}$ School of Data Science, The Chinese University of Hong Kong (Shenzhen), Shenzhen, China\\
    {\tt\small qhliu@sdu.edu.cn}
}

\begin{document}
\maketitle
\begin{abstract}
Eye-based emotion recognition enables eyewear devices to perceive users’ emotional states and support emotion-aware interaction. However, deploying such functionality on their resource-limited embedded hardware remains challenging. Time-to-first-spike (TTFS)-coded spiking neural networks (SNNs) offer a promising solution due to their extremely sparse and energy-efficient computation, where each neuron emits at most one binary spike. While prior works have primarily focused on improving TTFS SNN training algorithms, the role of network architecture has been largely overlooked. This is particularly critical, as spike timing in TTFS SNNs is tightly coupled with architectural design, and eye-based emotion recognition requires compact yet highly efficient networks. In this paper, we propose TNAS-ER, the first neural architecture search (NAS) framework tailored to TTFS SNNs for eye-based emotion recognition. TNAS-ER presents a novel ANN-assisted search strategy that leverages a ReLU-based ANN counterpart to guide architecture optimization and stabilize training of the TTFS SNN. TNAS-ER employs an evolutionary algorithm, with weighted and unweighted average recall jointly defined as fitness objectives for emotion recognition. Extensive experiments demonstrate that TNAS-ER achieves high recognition performance with significantly improved efficiency. Furthermore, we evaluate TNAS-ER on a neuromorphic hardware, confirming its superior energy efficiency and strong potential for real-world applications.
\end{abstract}
\section{Introduction}
\label{sec:intro}

Eye-based emotion recognition has gained increasing attention for its non-invasive, privacy-preserving, and flexible nature compared to facial or EEG emotion recognition \cite{wan2025eye}. Recent approaches employ event cameras to avoid motion blur and support a wider dynamic range than conventional RGB cameras \cite{gallego2020event}, enabling a precise capture of eye regions under challenging motion and lighting conditions. These advantages make eye-based emotion recognition particularly promising for wearable applications, such as immersive VR/AR interactions and emotion-aware mobile services \cite{hickson2019eyemotion,wu2020emo,cowie2001emotion}.

Despite its potential, deploying eye-based emotion recognition models on embedded eyewear platforms remains challenging. As these devices are compact and battery-powered, their memory, computational and energy resources are often limited \cite{wu2020emo}, necessitating recognition systems that are both accurate and highly efficient.

\begin{figure}
    \centering
    \includegraphics[width=0.85\linewidth]{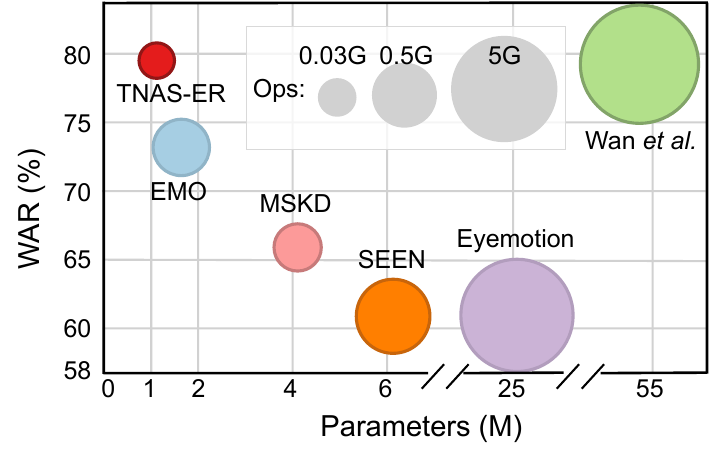}
    \caption{Comparison of TNAS-ER with state-of-the-art eye-based emotion recognition models on the SEE dataset, in terms of Weighted Average Recall (WAR), the number of parameters and operations. Circle size reflects the FLOPs (or SynOps for SNNs).}
    \label{fig:comparison}
\end{figure}

Time-to-first-spike-coded (TTFS) spiking neural networks (SNNs) offer a promising direction for efficient computing. SNNs are brain-inspired models that process information through discrete spike events, enabling event-driven and energy-efficient processing \cite{maass1997networks,roy2019towards}. In TTFS SNNs, each neuron emits at most one spike, resulting in extremely sparse activity and significantly reduced computation \cite{kheradpisheh2020temporal, yang2023lc}. This property makes TTFS SNNs well-suited for efficient inference on resource-constrained devices. 

 Existing research has primarily focused on developing training algorithms to improve the performance of TTFS SNNs \cite{wei2023temporal,yang2023lc,stanojevic2024high,zhaottfsformer}, while the role of network architecture in determining both accuracy and computational efficiency has been largely overlooked. However, architectural design is particularly critical. First, in TTFS SNNs, spike timing is tightly coupled with network architecture, making feature representations and overall performance sensitive to architecture choice. Second, practical wearable emotion recognition systems impose strict constraints on model size and energy consumption, requiring networks that are not only accurate but also highly compact and efficient. These considerations highlight the importance of exploring architecture design strategies tailored to TTFS SNNs for eye-based emotion recognition. Nevertheless, TTFS SNNs remain difficult to design, as their extremely sparse and timing-dependent spike activity leads to unstable gradients and challenging optimization.

This paper proposes TNAS-ER, a novel neural architecture search (NAS) framework tailored to TTFS SNNs for eye-based emotion recognition. TNAS-ER leverages a ReLU-based ANN (hereafter ReLU ANN) counterpart to stabilize training, guide architecture optimization, and enhance performance, effectively mitigating the difficulty of directly applying TTFS SNNs in NAS. By analyzing how ReLU ANN and TTFS SNN collaborate at each stage of the search, an ANN-assisted search strategy that bridges the two domains is introduced. TNAS-ER employs an evolutionary algorithm, with weighted and unweighted average recall jointly defined as fitness objectives for emotion recognition. Extensive experiments on two publicly available event-based single-eye emotion recognition datasets demonstrate that TNAS-ER achieves high recognition performance with significantly improved efficiency (see Figure~\ref{fig:comparison}). Furthermore, we deploy TNAS-ER on the neuromorphic hardware YOSO \cite{yu2023ttfs,chu2020you}, and the overall system achieves a low latency of 48 ms and energy consumption of 9.0 mJ, confirming its superior energy efficiency and strong potential for real-world applications.

The main contributions of this work are summarized as follows:
\begin{itemize}
\item We present the first exploration of TTFS SNNs for eye-based emotion recognition, achieving state-of-the-art accuracy with over 1.6$\times$, 5.6$\times$ and 5.8 $\times$ reductions in parameters, operations and latency, respectively, compared to existing ANN and SNN baselines.
\item We develop TNAS-ER, the first NAS framework specifically designed for TTFS SNNs, which automatically discovers efficient and high-performing architectures and outperforms manually designed and randomly searched architectures.
\item We introduce an ANN-assisted search strategy that leverages the mapping between TTFS SNNs and ReLU ANNs to effectively stabilize training and guide architecture optimization.
\end{itemize}

\section{Related Work}
\label{sec:relatedwork}

\subsection{Eye-based Emotion Recognition}
Eye-based emotion recognition has emerged as a promising alternative to facial or EEG-based approaches \cite{hickson2019eyemotion}. Compared with EEG-based methods, it offers a more convenient and comfortable user experience, and unlike facial analysis, it remains effective when the face is partially occluded (e.g., by masks or VR headsets) or when privacy concerns restrict full-face data usage.
Recent studies have shifted toward single-eye observations to overcome camera synchronization and bandwidth constraints in binocular setups~\cite{wu2020emo}.
To ensure reliable perception under challenging motion and lighting conditions, Zhang \textit{et al.} \cite{zhang2023blink} employed event cameras to capture the eye region and proposed an ANN-SNN hybrid network for joint frame–event processing. Han \textit{et al.} \cite{han2025hierarchical} designed a hierarchical event–RGB fusion framework for multi-scale semantics integration. Although these multimodal methods achieve high accuracy, they require handling both frame and event streams, resulting in large and computationally intensive networks. To improve efficiency, Wan \textit{et al.} \cite{wan2025eye} proposed a sparse transformer for event-based single-eye emotion recognition; however, the model still relies on a relatively large architecture.
Wang \textit{et al.} \cite{wang2024apprenticeship} distilled a large ANN–SNN teacher network into a unimodal SNN, reducing computation cost but depending on manually designed architectures that limit adaptability and efficiency.

\subsection{Spiking Neural Networks with TTFS Coding}
SNNs are brain-inspired models that transmit information through discrete spikes, mimicking biological communication \cite{roy2019towards}. Unlike ANNs, they operate in an event-driven manner, activating computations only when spikes occur. TTFS SNNs encode information in spike timing, allowing each neuron to fire at most once and thereby achieving extremely sparse and energy-efficient computation. Early works such as Rueckauer \textit{et al.} \cite{rueckauer2018conversion} converted ANNs to TTFS SNNs but were limited to shallow networks. Zhang \textit{et al.} \cite{zhang2019tdsnn} extended the conversion to deeper networks with reverse coding, and Park \textit{et al.} \cite{park2020t2fsnn} further improved it using dynamic thresholds and dendritic mechanisms. Yang \textit{et al.} \cite{yang2023lc} later proposed a near-lossless conversion strategy addressing temporal distortions. Zhao \textit{et al.} \cite{zhaottfsformer} proposed a conversion method for transformer architecture with TTFS-coding. In addition to conversion-based methods, Wei \textit{et al.} \cite{wei2023temporal} enabled direct TTFS training from pretrained ANNs. Stanojevic \textit{et al.} \cite{stanojevic2024high} established an identity mapping between ReLU networks and TTFS SNNs, enabling both conversion-based and direct training of TTFS models. Despite these advances, their architectural design remains unexplored. Given the crucial role of architecture in determining accuracy and efficiency \cite{na2022autosnn,yan2024efficient} and the intrinsic coupling between spike timing and architectural design in TTFS SNNs, an exploration of TTFS network architecture design is essential.

\subsection{Neural Architecture Search}
Neural Architecture Search (NAS) aims to automatically discover high-performing neural network architectures. Existing NAS methods generally fall into three categories: reinforcement learning (RL)-based \cite{bello2017neural,zoph2016neural,jaafra2019reinforcement}, gradient-based \cite{caiproxylessnas,liudarts,wu2019fbnet}, and evolutionary computation (EC)-based approaches \cite{liu2021survey,zhu2021real,pan2024brain}. Since RL-based NAS typically incurs high computational costs and large action spaces \cite{liu2021survey}, most SNN-based NAS frameworks adopt gradient-based or EC-based approaches. Che \textit{et al.} \cite{che2022differentiable} adapted DARTS to develop the first differentiable hierarchical search for SNNs, and Liu \textit{et al.} \cite{liu2024lite} extended it by incorporating spatial and temporal compression. Yan \textit{et al.} \cite{yan2024efficient} proposed a branchless supernet for gradient-based multi-objective optimization. However, gradient-based NAS relies on stable gradients and strong learning capabilities, which are difficult to satisfy in TTFS SNNs due to their sparse and timing-dependent nature. Na \textit{et al.} \cite{na2022autosnn} proposed an EC-based NAS framework with an effective SNN search space and backbone design. Che \textit{et al.} \cite{che2024auto} introduced evolutionary SNN neurons for transformer architecture search. Pan \textit{et al.} \cite{pan2024brain} presented a multi-scale evolutionary search inspired by brain topologies. Building on these advances, we adopt EC-based NAS for TTFS SNNs, which avoids bi-level optimization instability and enables a more reliable search process.

\section{TNAS-ER Framework}
\subsection{Preliminary}
\begin{figure}[!b]
    \centering
    \includegraphics[width=1.0\linewidth]{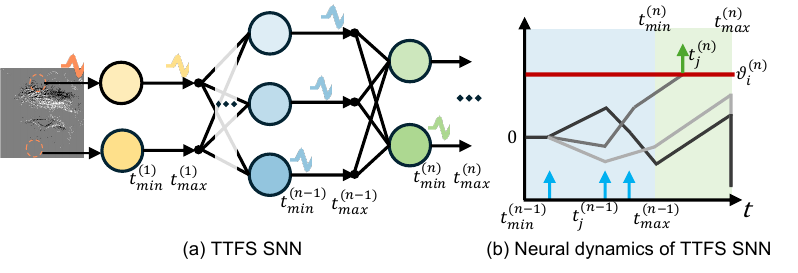}
    \caption{TTFS SNN and its neural dynamics.}
    \label{fig:TTFS}
\end{figure}
\textbf{TTFS Spiking Neurons:}
In TTFS SNNs, information is encoded by spike timing. Given the spike timing $t_j^{n-1}$ of presynaptic neurons $j$, the membrane potential $V_i^{n}$ of neuron $i$ in layer $n$ updates as \cite{stanojevic2024high}:
\begin{equation}
\tau_c \frac{dV_i^{(n)}}{dt} =
\begin{cases}
\displaystyle\sum_j W_{ij}^{n} H\!\left(t - t_j^{n-1}\right),
\hfill  t < t_{\min}^{n} \\
1, \hfill  t_{\min}^{n} \le t \le t_{\max}^{n}
\end{cases}
\label{equ:ttfs}
\end{equation}
where $W_{ij}^{(n)}$ denotes the synapse weight, $\tau_c$ is the time constant, and $H$ denotes the Heaviside function. $t_{\min}^{(n)}$ and $t_{\max}^{(n)}$ define the temporal bounds of the two phases, with $t_{min}^{(n)} = t_{max}^{(n-1)}$. As illustrated in Figure~\ref{fig:TTFS}, the membrane potential integrates incoming spikes before $t_{\min}^{(n)}$, then increases linearly with a slope of 1, defining the $B$1 model~\cite{stanojevic2024high}. A neuron fires when
$V_{i}^{(n)} \geq \vartheta_{i}^{(n)}$ and then enters a refractory period, ensuring that each neuron fires at most once. 
\begin{figure*}
    \centering
    \includegraphics[width=1\linewidth]{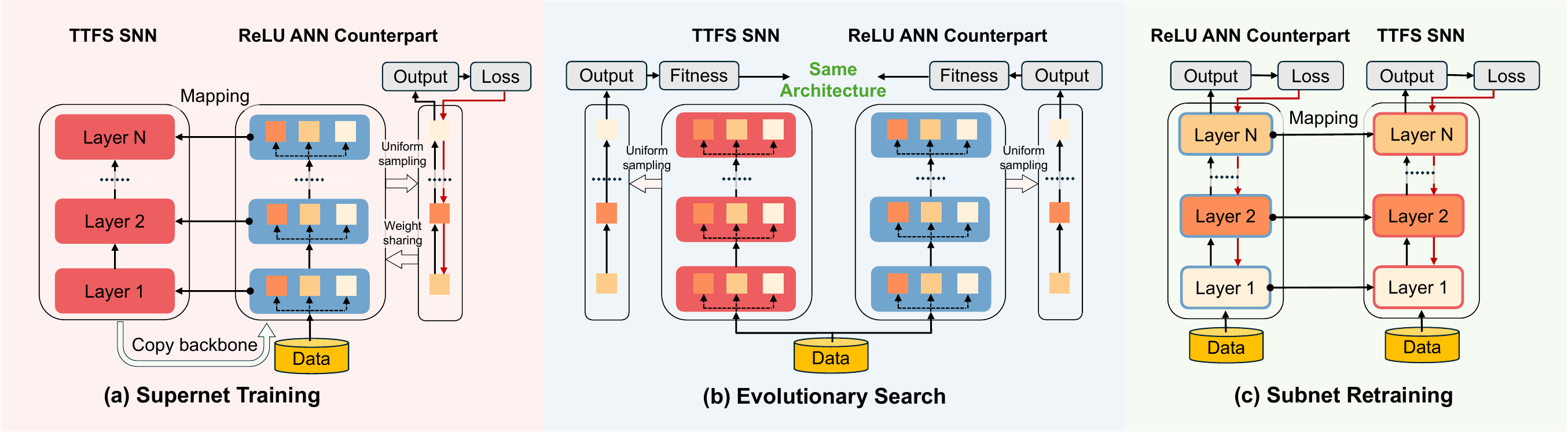}
    \caption{ANN-assisted TTFS SNN search strategy in TNAS-ER: (a) Supernet training with a ReLU ANN counterpart sharing the same backbone as the TTFS SNN and can be mapped back to the spiking domain; (b) Evolutionary search can be performed on either the TTFS SNN or the ANN counterpart, yielding identical high-performing architectures; (c) Subnet retraining in the ANN domain, followed by transfer to the TTFS SNN and SNN fine-tuning. }
    \label{fig:ANN-assistant}
\end{figure*}
\noindent \textbf{Mapping Between TTFS SNN and ReLU ANN:}
Under the above model, a bidirectional mapping can be established between TTFS SNNs and their counterpart ReLU ANNs with the same architecture \cite{stanojevic2024high}. The firing time is linearly related to the ReLU activation $x_i^{(n)}$:
\begin{equation}
t_i^{(n)} = t_{max}^{(n)} - x_i^{(n)}
\end{equation}
The temporal window is defined as:
\begin{equation}
t_{max}^{(n)} = t_{min}^{(n)} + (1 + \zeta) \cdot X_{max}^{(n)}
\end{equation}
where $\zeta$ is a scaling factor, ensuring that the time window is large enough to encode the maximum possible activation value from ReLU ANNs.
Given the synaptic weights $W_{ij}^n$ and firing thresholds $\vartheta_i^{n}$ of the SNN, the corresponding weight $w$ and bias $b$ of the counterpart ReLU network are defined as \cite{stanojevic2024high}:
\begin{equation}
w_{ij}^{(n)} = \mathcal{M}({W_{ij}^{(n)}}) = W_{ij}^{(n)}, \quad
b_i^{(n)} = 
-\vartheta_i^{(n)}
+ \frac{t_{\max}^{(n)} - t_{\min}^{(n)}}{\tau_c}
\end{equation}
where $\mathcal{M}$ is a function that maps the TTFS SNN weights $W_{ij}^{(n)}$ to the ReLU ANN weights $w_{ij}^{(n)}$.
Under this mapping, two networks can produce identical logits and losses for the same input. Tanojevic \textit{et al.} further demonstrated that a TTFS SNN and its ReLU ANN counterpart exhibit identical weight updates and training trajectories \cite{stanojevic2024high}:
 \begin{equation}
 \begin{aligned}
\delta w_{ij}^{(n)} &=\mathcal{M}(W_{ij}^{(n)}-\eta\frac{\mathrm{d}\mathcal{L}}{\mathrm{d}W_{ij}^{(n)}})-\mathcal{M}(W_{ij}^{(n)})\\
&= 
-\eta\frac{\mathrm{d}\mathcal{L}}{\mathrm{d}W_{ij}^{(n)}} = \Delta W_{ij}^{(n)}
\end{aligned}
\end{equation}
 This mapping between ReLU ANN and TTFS SNN allows TTFS SNN to inherit the learning capabilities of ANNs to enhance the performance, forming the foundation for our architectural design.
 
\subsection{ANN-Assisted TTFS SNN Search}
Neural architecture design is crucial for enhancing the efficiency and capability of SNNs \cite{na2022autosnn, liu2024lite}. However, TTFS SNNs present unique challenges due to their single-spike temporal coding and layer-wise time scheduling, which result in unstable gradients and high 
sensitivity to architectural configurations \cite{stanojevic2024high}. To address this challenge, TNAS-ER proposes an ANN-assisted search strategy (see Figure \ref{fig:ANN-assistant}). Instead of relying solely on TTFS SNN learning, we leverage a ReLU ANN counterpart that shares a bidirectional mapping with the TTFS SNN to assist the search process. The framework consists of three stages: supernet training, evolutionary search, and subnet retraining with fine-tuning. We describe how  TTFS SNN and the ANN counterpart collaborate at each stage below.

\begin{figure}[!t]
    \centering
    \includegraphics[width=1.0\linewidth]{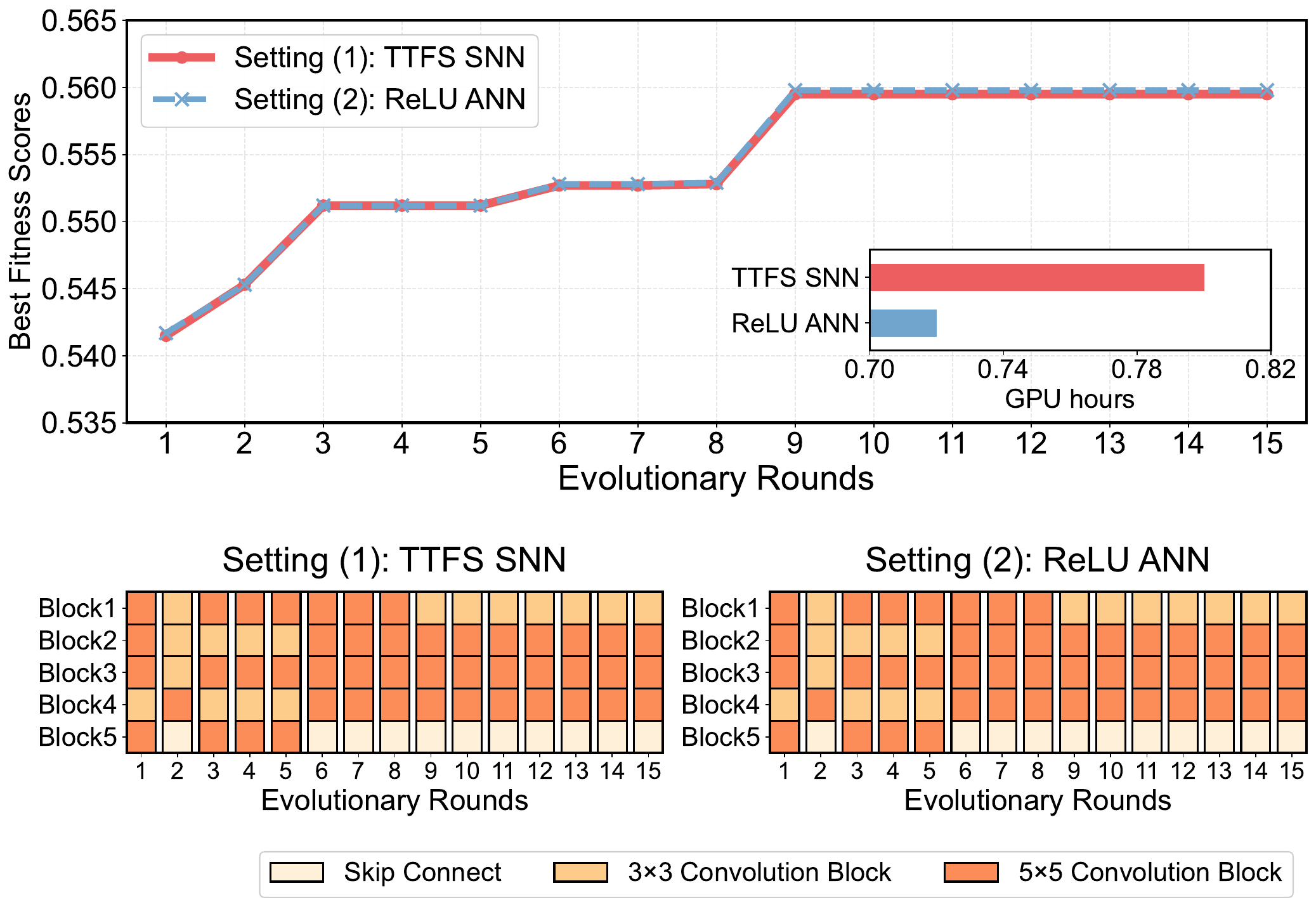}
    \caption{Evolutionary search quality comparison between TTFS SNN and its counterpart ANN: Top - fitness curves and search time, Bottom - best architecture evolution.}
    \label{fig:method-search}
\end{figure}

\noindent\textbf{Supernet training.}
We first construct an SNN supernet composed of searchable blocks, each containing multiple candidate operations (see Section~\ref{ss}). During training, we adopt a uniform sampling strategy to activate different candidate paths, enabling efficient exploration of the search space. However, directly training a TTFS SNN supernet is highly unstable, as different candidate operations may yield inconsistent $t_{\max}^{(n)}$ values, causing temporal misalignment within one layer. Moreover, NAS supernets typically include skip operations to explore variable network depths, yet in TTFS SNNs, such connections disrupt strict layer-wise temporal scheduling. 

To address these issues, we introduce the ReLU ANN counterpart that shares the same backbone architecture as the TTFS SNN and conduct the supernet training in the ANN domain. Unlike TTFS SNN, ReLU ANN operates on continuous activations, avoiding temporal dependencies and enabling stable gradient-based optimization. This makes it more suitable for training large and flexible supernets with diverse candidate operations. After training, the learned parameters can be transferred to the TTFS SNN via the mapping. In this way, the ANN serves as an optimization surrogate that provides stable training signals while preserving compatibility with the TTFS SNN. Overall, this ANN-assisted training scheme eliminates temporal inconsistencies and provides better-initialized weights for candidate blocks, thereby improving search reliability and facilitating the discovery of high-performing architectures.

\noindent\textbf{Evolutionary search.}
During the evolutionary search, candidate subnets are iteratively sampled, evaluated, and evolved to discover high-performing architectures. Each candidate subnet is treated as an individual, whose fitness is evaluated through forward inference using the inherited supernet weights. No parameter learning is involved at this stage. Higher-performing candidates generate new ones via mutation and crossover, progressively guiding the search toward optimal designs.

We consider two settings for architecture discovery: (1) transferring the trained ReLU ANN supernet to TTFS SNN and conducting the search based on SNN, and (2) performing the search directly on the trained ReLU ANN counterpart. In both settings, the network accuracy obtained from the forward pass is used as the fitness score. As shown in Figure~\ref{fig:method-search}, using a search space with five searchable blocks (details in Section~3.3), the two settings yield nearly identical fitness curves and converge to the same architectures across all search rounds. This consistency arises from the mapping between TTFS SNNs and ReLU ANNs, which ensures consistent logits for the same inputs. Consequently, evolutionary search can be equivalently performed on either the ReLU ANN or the TTFS SNN. In practice, we perform the search in the ANN domain (setting (2)), as evaluating TTFS SNNs requires additional configuration of $t_{\max}$, resulting in longer search time. 
\begin{figure}[!t]
    \centering
    \includegraphics[width=0.8\linewidth]{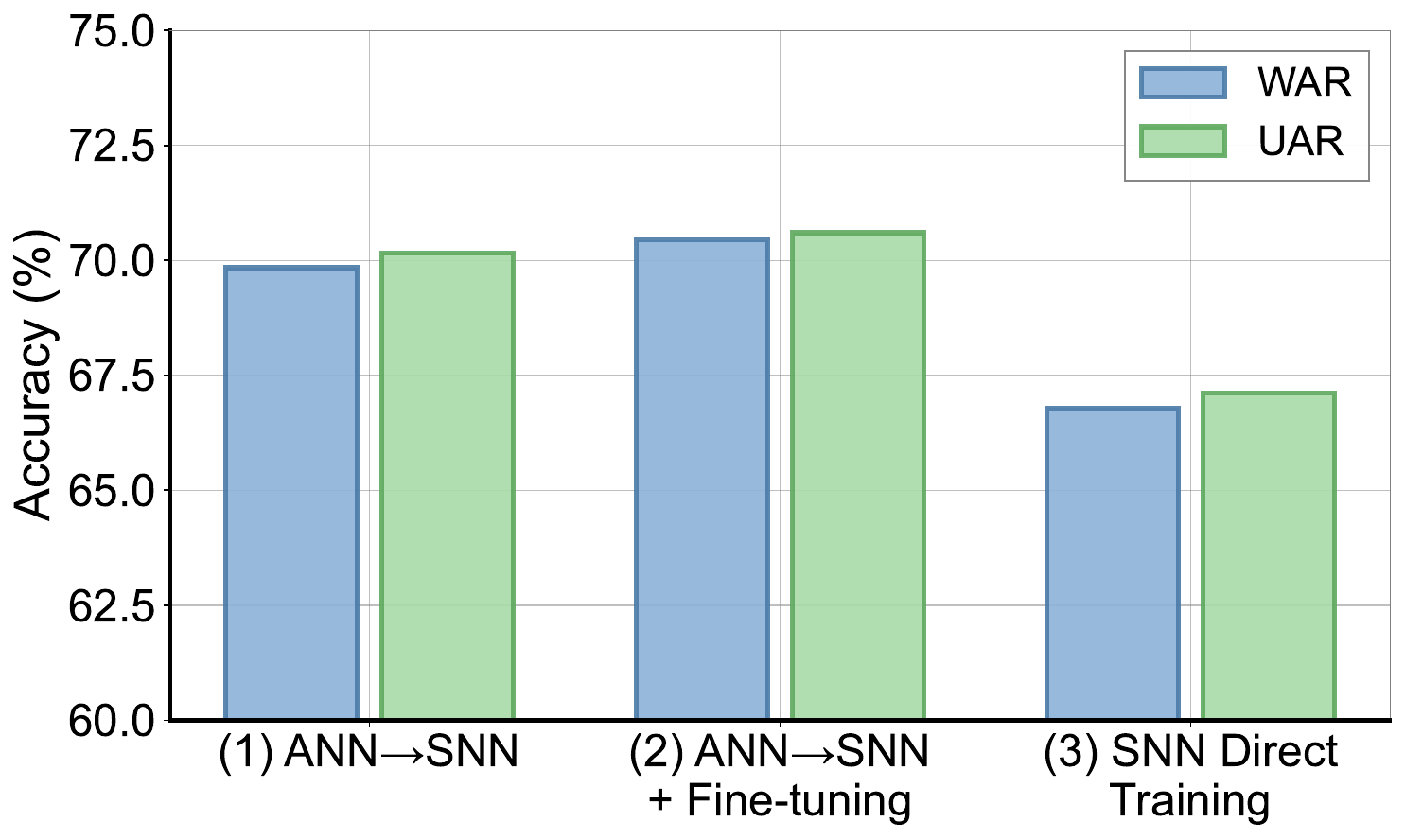}
    \caption{Subnet retrain quality comparison across different strategies.}
    \label{fig:method-retrain}
\end{figure}
\noindent\textbf{Subnet retraining with fine-tuning.}
For the selected architecture, we explore three retraining settings: (1) retraining in the ANN domain and transferring to the TTFS SNN via the mapping,
(2) performing (1) followed by SNN fine-tuning, and (3) directly retraining the SNN from scratch. Using a backbone where all searchable blocks adopt 3$\times$3 convolution kernels, we observe in Figure~\ref{fig:method-retrain} that ANN-assisted settings (1) and (2) outperform direct SNN retraining, highlighting the superior learning capability provided by the ANN counterpart. Although (1) and (3) share theoretically identical training trajectories, their accuracies diverge because this equivalence holds only in the absence of Batch Normalization (BN). In practice, (1) trains the ReLU ANN with BN, fuses BN into the weights, and transfers them to the TTFS SNN, yielding higher accuracy than (3), which trains without BN. Furthermore, setting (2) fine-tunes the trained model to better adapt to spiking dynamics while supporting adaptation to deployment requirements for target hardware.

It is worth noting that although TNAS-ER leverages the ANN counterpart to assist the search, the search objective, search space, and performance evaluation are all defined within the TTFS SNN domain. The ReLU ANN merely serves as a differentiable proxy, enabling effective network training that would otherwise be unstable to apply directly in TTFS SNNs. Both the supernet and the subnet are trained using the cross-entropy loss. The pseudocode of TNAS-ER is provided in Algorithm~\ref{alg:TNAS_ER}.
\begin{figure}
    \centering
    \includegraphics[width=0.9\linewidth]{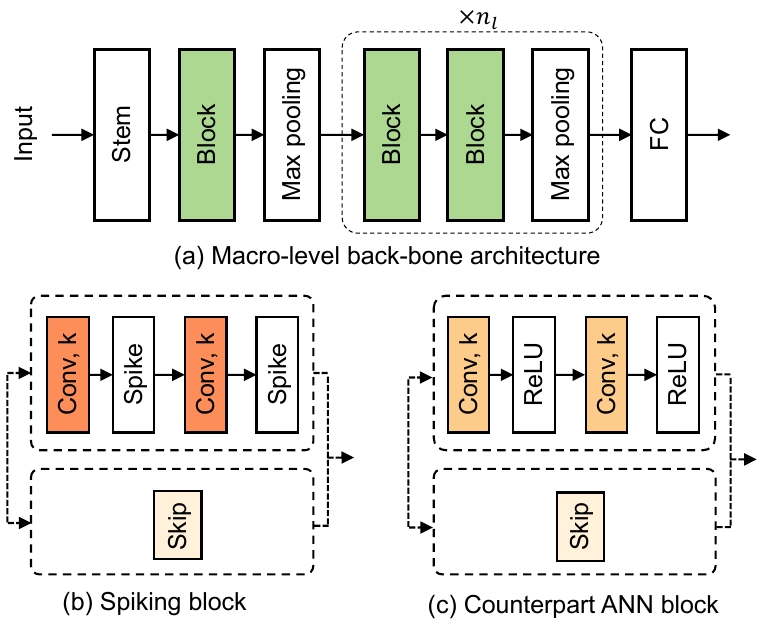}
    \caption{Hierarchical search space. Upper: Macro-level backbone architecture, composed of searchable blocks. Bottom: Within one searchable block, the micro-level spiking block and its ANN counterpart.}
    \label{fig:searchspace}
\end{figure}
\subsection{Search Space and Search Algorithm}
\label{ss}
\noindent\textbf{Search Space.} 
We adopt an effective hierarchical search space inspired by prior SNN NAS studies~\cite{na2022autosnn}. As shown in Figure~\ref{fig:searchspace}, it consists of a macro-level backbone and a micro-level spiking block. At the macro level, the backbone includes a stem, multiple searchable blocks separated by max-pooling, and a fully connected head. At the micro level, each searchable block contains convolution blocks with different $k\times k$ kernel sizes or a skip operation. Residual shortcuts are excluded since TTFS SNNs process layerwise, non-overlapping temporal windows, where cross-layer links would misalign spike timing. Both skip operations and residual shortcuts introduce identity-like connections. However, skip operations are temporary during search, and any resulting inconsistency can be mitigated via ANN-assisted training, whereas residual shortcuts introduce permanent cross-path aggregation that disrupts the layer-wise temporal structure of TTFS SNNs. Each spiking convolution block comprises two convolution layers with searched kernel sizes interleaved with spiking activations. In the ANN counterpart, the activations are replaced by ReLU functions. BN layers can be applied during training but are fused into the weights when transferring to the TTFS SNN, ensuring compatibility with the mapping.

\begin{algorithm}[!t]
\caption{TNAS-ER Framework}
\label{alg:TNAS_ER}
\begin{algorithmic}[1]
\Statex \textbf{Supernet Training Phase}
\Require Training data $D_{\text{train}}$, supernet $\mathcal{S}_{\text{ANN}}$, max train iteration $\mathcal{T}_{train}$
\Ensure Trained supernet parameters $\Theta$
\For{$r = 1$ to $\mathcal{T}_{\text{train}}$} 
    \State Sample a subnet ($\mathcal{A}) \sim \text{Uniform}(\mathcal{S}_{\text{ANN}})$
    \State $\hat{y} \gets \mathcal{A}(D_{\text{train}}; \Theta)$
    \State \text{Compute loss and update the shared parameters $\Theta$}
\EndFor
\State \Return $\Theta$
\Statex \textbf{Evolutionary Search Phase}
\Require Trained supernet $\mathcal{S}$ with parameters $\Theta$, validation data $D_{\text{val}}$, evaluation pool size $N_{\text{eval}}$, elite pool size $N_{\text{top}}$, max search iteration $\mathcal{T}_{search}$
\Ensure Architecture $\mathcal{A}^*$ with the highest fitness
\State Initialize evaluation pool  $P_{\text{eval}}$
\For{$r = 1$ to $\mathcal{T}_{\text{search}}$} 
    \State $P_{\text{top}} \gets \text{TopK}(P_{\text{eval}}, N_{\text{top}})$
    \State $P_{\text{eval}} \gets \text{Mutation}(P_{\text{top}}) \cup \text{Crossover}(P_{\text{top}}) $
    \State $P_{\text{eval}} \gets P_{\text{eval}} \cup \text{Uniform}(\mathcal{S}, N_{\text{eval}} - |P_{\text{eval}}|)$
    \State fitness score $\gets \mathcal{A}(D_{\text{val}}; \Theta)$ for each $\mathcal{A}$ in $P_{\text{eval}}$
\EndFor
\State \Return Top-1 architecture $\mathcal{A}^*$ in $P_{\text{top}}$
\Statex \textbf{Subnet Retraining with Finetuning Phase}
\Require Searched architecture $\mathcal{A}^*$, trained supernet $\mathcal{S}_{ANN}$, $\mathcal{S}_{SNN}$ with parameters $\Theta$, training data $D_{\text{train}}$, max retrain and finetune iteration $\mathcal{T}_{retrain}$, $\mathcal{T}_{ft}$
\Ensure Trained SNN parameters $\theta_{\text{SNN}}$
\For{$r = 1$ to $\mathcal{T}_{\text{retrain}}$} 
    \State $y \gets \mathcal{A}^*(D_{\text{train}}; \theta)$
    \State \text{Compute loss and update parameters $\theta$}
\EndFor
\State Convert $\mathcal{A}^*(\theta)$ to TTFS SNN $\mathcal{A^*_{SNN}}(\theta_{SNN})$
\For{$r = 1$ to $\mathcal{T}_{\text{ft}}$} 
    \State $y \gets \mathcal{A}^*_{\text{SNN}}(D_{\text{train}}; \theta_{\text{SNN}})$
    \State \text{Compute loss and update $\theta_{\text{SNN}}$}
\EndFor
\State \Return $\theta_{\text{SNN}}$
\end{algorithmic}
\end{algorithm}

\begin{table*}[t]
\centering
\caption{Comparison of different methods on the \textbf{SEE} dataset. The best and the second-best results are highlighted with \uuline{number} and \underline{number}, respectively. The abbreviations are defined as Ha$\rightarrow$ Happiness, Sa $\rightarrow$ Sadness, An $\rightarrow$ Anger, Di $\rightarrow$ Disgust, Su $\rightarrow$ Surprise, Fe $\rightarrow$ Fear, Ne $\rightarrow$ Neutral, Nor $\rightarrow$ Normal, Over $\rightarrow$ Overexposure, Low $\rightarrow$ Low-light. “Hybrid” denotes architectures combining both SNN and ANN components.
}
\label{tab:comparison1}
\resizebox{\textwidth}{!}{%
\begin{tabular}{l | c | c | c c c c c c c | c c c c | c c}
\hline
\multirow{2}{*}{\textbf{Methods}} & \multirow{2}{*}{\textbf{Network}}&
\multirow{2}{*}{\textbf{Input region}} & 
\multicolumn{7}{c|}{\textbf{Accuracy for different emotions (\%)}} & 
\multicolumn{4}{c|}{\textbf{Accuracy for different lightings (\%)}} & 
\multicolumn{2}{c}{\textbf{Metrics (\%)}} \\
&& & Ha & Sa & An & Di & Su & Fe & Ne & Nor & Over & Low & HDR & WAR$\uparrow$ & UAR$\uparrow$ \\
\hline
Resnet18 + LSTM  & ANN & Face  & 37.0 & 81.2 & 38.9 & 44.7 & 39.0 & 48.5 & 56.4 & 53.2 & 49.5 & 50.4 & 40.8 & 48.6 & 49.4  \\
Resnet50 + GRU  & ANN      & Face &  31.9 & 70.1 & 42.0 & 48.0 & 49.2 & 46.0 & 58.7 & 52.1 & 51.6 & 51.6 & 41.1 & 49.3 & 49.4 \\
3D Resnet18~\cite{hara2018can}  & ANN & Face & 79.6 & \underline{93.7} & 72.9& \uuline{77.1} & 58.5 & 75.3 & 83.6 & 75.6 & 75.3 & 80.2 & 73.9 & 76.3 & 77.2 \\
R(2+1)D~\cite{tran2018closer}  & ANN    & Face & \underline{82.0} & 93.6 & 77.9 & 72.5 & 61.5 & 70.4 & \underline{84.1} & 76.6 & 76.2 & 79.6 & 73.1 & 76.4 & 77.4 \\
Former DFER~\cite{zhao2021former} & ANN & Face& 74.2 & 89.4 & \underline{80.0} & 68.1 & 61.7 & 72.5 & 74.5 & 75.3 & 74.3 & 74.0 & 71.3 & 73.8 & 74.3 \\
Eyemotion~\cite{hickson2019eyemotion} & ANN & Eye & 63.7 & 78.0 & 59.5 & 57.9 & 44.7 & 60.1 & 71.0 & 60.8 & 57.8 & 64.4 & 61.1 & 61.0 & 62.0 \\
EMO  \cite{wu2020emo}     & ANN    & Eye & 72.6 & 91.2 & 79.8 & 65.5 & 57.7 & \underline{77.8} & 76.6 & 73.3 & 74.0 & 77.8 & 69.6 & 73.7 & 74.5 \\
SEEN   \cite{zhang2023blink} & Hybrid&Eye & 58.5 & 76.9 & 67.4 & 57.1 & 50.7 & 60.9 & 64.7 & 61.8 & 64.1 & 65.9 & 55.1 & 61.8 & 62.3 \\
MSKD \cite{wang2024apprenticeship}  &SNN & Eye & 64.4 & 82.3 & 71.0 & 60.9 & 52.8 & 67.1 & 70.1 & 65.7 & 67.7 & 68.4 & 63.1 & 66.3 & 66.9 \\
Wan \textit{et al.} \cite{wan2025eye} & ANN & Eye  & \uuline{84.2} & {92.8} & {79.9} & \underline{76.7} & \underline{66.2} & {75.8} & \uuline{84.2} & \uuline{78.4} & \uuline{80.6} & \underline{83.1} & \underline{74.4} & \underline{79.2} & \underline{80.0} \\
\hline
\textbf{TNAS-ER}& SNN &Eye & 77.8 & \uuline{95.0} & \uuline{81.3} & 72.1 & \uuline{71.2} & \uuline{82.3} & 81.2 & \underline{78.1} & \underline{80.5} & \uuline{83.8} & \uuline{75.8} &\uuline{79.6}  & \uuline{80.1}  \\
\hline
\end{tabular}
}
\end{table*}

\noindent\textbf{Search Algorithm.} 
As shown in Figure \ref{fig:ANN-assistant}, TNAS-ER adopts a one-shot weight-sharing strategy based on the evolutionary algorithm. During supernet training, single-path uniform sampling is used to randomly sample subnets, train them, and share their updated weights back to the supernet. Afterward, the evolutionary algorithm is applied to explore the search space and identify the architecture with the highest fitness value.
We define the fitness value score of an architecture $\mathcal{A}$ as:
\begin{equation}
F(\mathcal{A}) = \frac{WAR + UAR}{2},
\end{equation}
where UAR (Unweighted Average Recall) $= \frac{1}{C} \sum_{i=1}^{C} \frac{TP_i}{TP_i + FN_i}$ measures the average recall across all emotion classes, treating each class equally regardless of frequency, and WAR (Weighted Average Recall) $= \frac{TP+TN}{TP + TN + FP + FN}$ reflects the overall recognition accuracy across all samples. Combining both metrics provides a balanced evaluation of recognition accuracy and robustness to class imbalance. 
The evolutionary process begins by randomly sampling $N_{eval}$ candidate architectures to form an evaluation pool $P_{eval}$. We compute the fitness score of each architecture in $P_{eval}$ and select the top $N_{top}$ candidates as the elite set $P_{top}$. From $P_{top}$, we generate a new set of $N_{eval}$ architectures through mutation and crossover, forming the new $P_{eval}$. We then combine $P_{eval}$ and $P_{top}$ and retain the top $N_{top}$ architectures to update the elite pool. This process iterates until convergence, yielding the architecture with the highest fitness score. 

\begin{table*}[t]
\centering
\caption{Comparison of different methods on the \textbf{DSEE} dataset. The best and the second best results are highlighted in \uuline{number} and \underline{number}, respectively. The abbreviations are defined as Ha$\rightarrow$ Happiness, Sa $\rightarrow$ Sadness, An $\rightarrow$ Anger, Di $\rightarrow$ Disgust, Su $\rightarrow$ Surprise, Fe $\rightarrow$ Fear, Ne $\rightarrow$ Neutral, Nor $\rightarrow$ Normal, Over $\rightarrow$ Overexposure, Low $\rightarrow$ Low-light. “Hybrid” denotes architectures combining both SNN and ANN components.}
\label{tab:comparison2}
\resizebox{\textwidth}{!}{%
\begin{tabular}{l | c | c | c c c c c c c | c c c c | c c}
\hline
\multirow{2}{*}{\textbf{Methods}} & \multirow{2}{*}{\textbf{Network}} &
\multirow{2}{*}{\textbf{Input region}} & 
\multicolumn{7}{c|}{\textbf{Accuracy for different emotions (\%)}} & 
\multicolumn{4}{c|}{\textbf{Accuracy for different lightings (\%)}} & 
\multicolumn{2}{c}{\textbf{Metrics (\%)}} \\
&& & Ha & Sa & An & Di & Su & Fe & Ne & Nor & Over & Low & HDR & WAR$\uparrow$ & UAR$\uparrow$ \\
\hline
R(2+1)D~\cite{tran2018closer}    &ANN   & Face& 68.5 & 67.0 & 72.0 & \underline{75.0} & {68.7} & \uuline{62.4} & 73.0 & {68.9} & 66.3 & \underline{68.4} & {\underline{78.7}} & 69.4 & 69.5 \\
Former DFER~\cite{zhao2021former} &ANN &Face& 65.4 & 68.6 & 71.4 & 70.9 & 67.7 & \underline{57.0} & 77.6 & 68.3 & 66.4 & 66.2 & 70.9 & 67.9 & 68.4 \\
Eyemotion \cite{hickson2019eyemotion}& ANN & Eye & 63.0 & 50.2 & 57.3 & 52.2 & 48.9 & 56.6 & 61.8 & 55.5 & 49.8 & 54.0 & 64.5 & 55.2 & 55.7 \\
EMO  \cite{wu2020emo} &ANN        & Eye & 57.7 & 61.0 & 67.2 & 61.1 & 61.0 & 48.0 & 71.8 & 60.1 & 58.0 & 61.7 & 65.1 & 60.5 & 61.1 \\
SEEN   \cite{zhang2023blink} &Hybrid& Eye & 54.8 & 55.5 & 62.6 & 58.8 & 63.1 & 47.6 & 69.8 & 58.5 & 59.0 & 56.6 & 58.5 & 58.3 & 58.9 \\
MSKD  \cite{wang2024apprenticeship} &SNN& Eye & 61.3 & 59.3 & 65.6 & 60.9 & 63.3 & 47.4 & 67.1 & 60.1 & 59.1 & 61.4 & 63.0 & 60.4 & 60.7\\
Wan \textit{et al.} \cite{wan2025eye}&ANN& Eye & {\uuline{75.1}} & {\underline{70.6}} & \underline{72.9} & {\uuline{76.1}} & {\underline{73.0}} & 56.2 & {\uuline{80.6}} & {\uuline{72.3}} & \underline{69.4} & 67.8 & {77.8} & {\underline{71.6}} & {\underline{72.1}} \\
\hline
\textbf{TNAS-ER}& SNN &Eye & \underline{70.4} & \uuline{70.7} & \uuline{78.2} & 74.5 & \uuline{74.7} & \uuline{62.4} & \underline{79.3} & \underline{72.1} & \uuline{70.2} & \uuline{72.2} & \uuline{79.7} & \uuline{72.6} & \uuline{72.9} \\
\hline
\end{tabular}
}
\end{table*}

\section{Experiments}
\subsection{Datasets}
We evaluate the proposed TNAS-ER method on two publicly available single-eye event-based emotion datasets: SEE and DSEE. 

The SEE dataset \cite{zhang2023blink} was collected using a DAVIS346 event camera mounted on a helmet in front of the participant’s right eye to capture eye-region activity. It contains 71.5 minutes of recordings from 111 volunteers. Following prior works \cite{zhang2023blink,wang2024apprenticeship}, the dataset is split into 1,638 training samples and 767 testing samples.
SEE covers seven emotion categories (anger, happiness,
sadness, surprise, fear, disgust, and neutral) under four
lighting conditions (normal, high dynamic range (HDR), low-light, and overexposure).

The DSEE dataset \cite{wang2024apprenticeship} extends SEE by incorporating additional samples converted from five conventional
video-based emotion datasets \cite{lucey2010extended,aifanti2010mug,pantic2005web,zhao2011facial,van2011moving} using the video-to-event simulator~\cite{hu2021v2e}. It includes 6,235 samples from 394 subjects, covering a wider range of ages and ethnic backgrounds. This increased diversity makes DSEE a more comprehensive and challenging benchmark for evaluating event-based emotion recognition methods. Following \cite{wang2024apprenticeship}, the dataset is split into 4362 training samples and 1873 testing samples.

For both datasets, the training samples are further divided into two subsets: 80\% for supernet training ($\mathcal{D}_{train}$) and the remaining 20\% for evolutionary search ($\mathcal{D}_{val}$). During the retraining stage, all training samples are used. The testing samples are strictly held out and not used at any stage of the search process.

\subsection{Experimental Settings}
\noindent\textbf{Implementation Details.}
All experiments are implemented in PyTorch and conducted on an NVIDIA GeForce RTX 4090 GPU (24 GB). The learning rate is set to 5e-3, 5e-4, and 5e-5 for supernet training, retraining, and fine-tuning, respectively. The supernet and subnet are trained for 600 epochs, the evolutionary search runs for 18 rounds, and the SNN fine-tuning is performed for 100 epochs. The networks are optimized using Adam, and the batch size is 96. Following prior works \cite{zhang2023blink, wang2024apprenticeship}, we adopt UAR and WAR to quantitatively evaluate emotion recognition performance, and all reported results are averaged over 20 inference runs. 

\noindent\textbf{Preprocessing.} Event data preprocessing follows prior works~\cite{zhang2023blink,wang2024apprenticeship,wan2025eye}. Events are first aggregated into event segments, and the E4-S3 setting is adopted, where Ex-Sy denotes $x$ event frames with a skip interval of $y$ (in 1/30-second units) between frames, yielding a total test duration of $(x + (x - 1) \times y)/30$ seconds. All methods are evaluated under this setting, with multiple trials using random start points to ensure robustness. The event frames are then concatenated into 4 channels and fed into the network.

\begin{table}[t]
	\centering
        \caption{Computational efficiency comparison of different eye-based emotion recognition methods. For ANN-based methods, OPs refer to FLOPs, while for SNN-based methods, OPs are measured as SynOPs.}
        \resizebox{\linewidth}{!}{
		\begin{tabular}{lccccc}
			\toprule
			\textbf{Methods} & \textbf{WAR} (\%) & \textbf{UAR} (\%) & \textbf{OPs} (G)& \textbf{Params} (M) &\textbf{Latency} (ms) \\
            \midrule
			Eyemotion & 61.0 & 62.0 & 5.73 & 25.13 &45 \\
			EMO & 73.7 &74.5 & 0.35 & 1.68 & 19 \\
			SEEN & 61.8& 62.3 & 0.73 & 6.08 & 9 \\
			MSKD & 66.3&66.9 & 0.17 & 4.04  & 6 \\
            Wan \textit{et al.} & 79.2 & 80.0 & 8.44 & 51.94 & 19\\
             \midrule
             \textbf{TNAS-ER} & 79.6 & 80.1& 0.03 & 1.04 & 1.03\\
			\bottomrule
		\end{tabular}}
		\label{tab:model_compare}
\end{table}

\noindent\textbf{Architecture}
The stem block consists of two convolutional layers with $3 \times 3$ kernel size. The searched backbone includes $n_l = 2$ macro-block groups, each containing two searchable blocks. The first convolution in the stem block takes 4 input channels and outputs 32 channels, and the channel width doubles after each max-pooling layer. During the search, $N_{eval}$ and $N_{top}$ are set to 12. The factor $\zeta$ in the temporal window is set to 0.5.

\subsection{Performance Comparison}
Table~\ref{tab:comparison1} and Table~\ref{tab:comparison2} compare the proposed TNAS-ER with state-of-the-art methods on the SEE and DSEE datasets, following the same experimental settings as prior works~\cite{wang2024apprenticeship, wan2025eye} for fair evaluation.

TNAS-ER consistently achieves the highest performance on both datasets, outperforming all facial and eye-region methods. On the SEE dataset, TNAS-ER reaches a WAR of 79.6\% and a UAR of 80.1\%, and achieves top scores in four of the seven emotion categories (sadness, anger, surprise, and fear) and in all four lighting conditions (normal, overexposure, low-light, and HDR). Notably, compared with other SNN-based approaches such as MSKD and SEEN, TNAS-ER achieves substantially stronger performance, with improvements exceeding 10\% in both WAR and UAR. A similar trend is observed on the DSEE dataset, where our TNAS-ER obtains a WAR of 72.6\% and a UAR of 72.9\%, outperforming the best-performing baseline \cite{wan2025eye} by 1.0\% in WAR and 0.8\% in UAR. Furthermore, TNAS-ER achieves top scores across six of seven emotion categories (happiness, sadness, anger, surprise, fear, and neutral) and all lighting conditions. Since DSEE includes more subjects with a broader range of ages and ethnic backgrounds, the larger performance gain compared to SEE highlights the strong generalization and robustness of TNAS-ER across diverse and challenging scenarios. 

\subsection{Computational Efficiency}
This section evaluates the computational efficiency of the proposed TNAS-ER, which is essential for practical deployment. We compare TNAS-ER with existing eye-based approaches on the SEE dataset in terms of recognition performance (WAR and UAR) and computational cost (the number of operations, parameters, and inference latency).

As shown in Table~\ref{tab:model_compare}, TNAS-ER achieves the best balance between accuracy and efficiency. It surpasses all baseline methods with the highest recognition performance, while reducing operations, parameters, and latency by over 5.6$\times$, 1.6$\times$, and 5.8$\times$, respectively. Compared to the most accurate baseline Wan \textit{et al.}~\cite{wan2025eye}, TNAS-ER reduces the number of operations by over 280$\times$, parameters by nearly 50$\times$ and inference latency by over 17$\times$. Even when compared to the lightweight EMO model, TNAS-ER delivers nearly 6\% higher accuracy with 11$\times$ fewer operations, 1.6$\times$ fewer parameters and 17$\times$ lower latency. We further observe that both SNN-based methods, MSKD and ours, exhibit substantially fewer operations than ANN-based models, confirming that the intrinsic sparsity of SNNs contributes to improved computational efficiency. Owing to its compact architecture and TTFS coding, TNAS-ER achieves lower SynOPs than MSKD. These results collectively highlight the superior effectiveness and efficiency of TNAS-ER, making it particularly suitable for low-power, real-world deployments.

\begin{figure}
    \centering
    \includegraphics[width=1.0\linewidth]{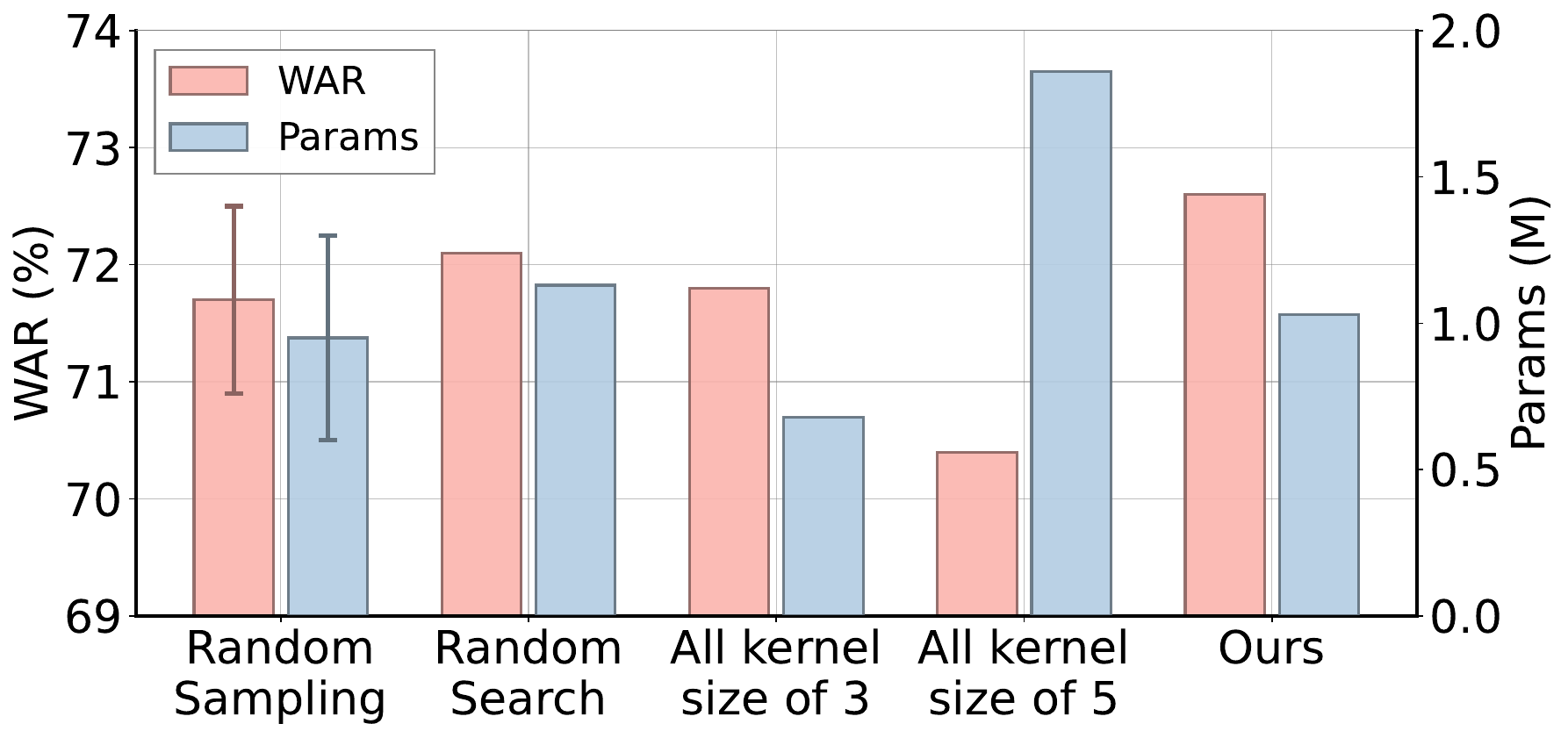}
    \caption{Performance comparison with SNNs obtained from different settings.}
    \label{fig:ablation}
\end{figure}
\subsection{Effectiveness of the Searched Architecture}
In this section, we conduct comparative experiments on the more challenging DSEE dataset to validate the effectiveness of the architecture discovered by TNAS-ER.

\noindent\textbf{Comparison with Random Sampling.}
We first randomly sample 10 architectures from the search space, where each searchable block independently selects a convolution block with a kernel size of $3 \times 3$ or $5 \times 5$, or a skip connection. Each sampled architecture is trained and evaluated independently, and the reported WAR and UAR values are averaged among 10 architectures.
As shown in Figure~\ref{fig:ablation}, the architectures obtained through random sampling achieve lower performance than TNAS-ER, highlighting the superiority of the searched architecture.

\noindent\textbf{Comparison with Random Search.}
We implement random search following \cite{na2022autosnn} by selecting the architecture with the highest fitness scores among 100 randomly sampled architectures based on the trained supernet. As shown in Figure~\ref{fig:ablation}, this approach yields better performance than random sampling but still falls short of TNAS-ER.
The result indicates that while fitness-based selection helps identify stronger TTFS SNN architectures, the evolutionary strategy in TNAS-ER further enhances search effectiveness.

\noindent\textbf{Comparison with Fixed Backbones.}
We further compare our searched architecture with fixed-backbone architectures, where all searchable blocks are replaced with convolution blocks of uniform kernel sizes ($3 \times 3$ or $5 \times 5$). As shown in Figure~\ref{fig:ablation}, both architectures with fixed configurations perform worse than the architecture discovered by TNAS-ER. We observe that the architecture using $3 \times 3$ kernels outperforms that using $5 \times 5$ kernels, indicating that a larger network does not necessarily lead to higher performance. These results demonstrate that the proposed evolutionary search can automatically discover an optimal balance between network capacity and efficiency, yielding compact yet high-performing architectures beyond what manual design can achieve.

\noindent\textbf{Architecture Transferability.}
We evaluate the transferability of the searched architecture to examine its adaptability across different datasets. Transferability is particularly important when users need to rapidly deploy an existing architecture to new tasks, avoiding the cost of searching from scratch. Since the DSEE dataset contains a broader range of samples, we first search for the optimal architecture on SEE and then train and test this architecture on DSEE. The transferred model achieves 72.1\% WAR and 72.3\% UAR, with less than a 1\% performance drop compared to the architecture searched directly on DSEE. Remarkably, it attains competitive results with Wan et al.~\cite{wan2025eye} while using only 1.04 M parameters, compared to their 51 M, demonstrating strong generalization and efficiency of the searched architecture.

\subsection{Evaluation on Neuromorphic Hardware}
\begin{figure}
    \centering
    \includegraphics[width=1\linewidth]{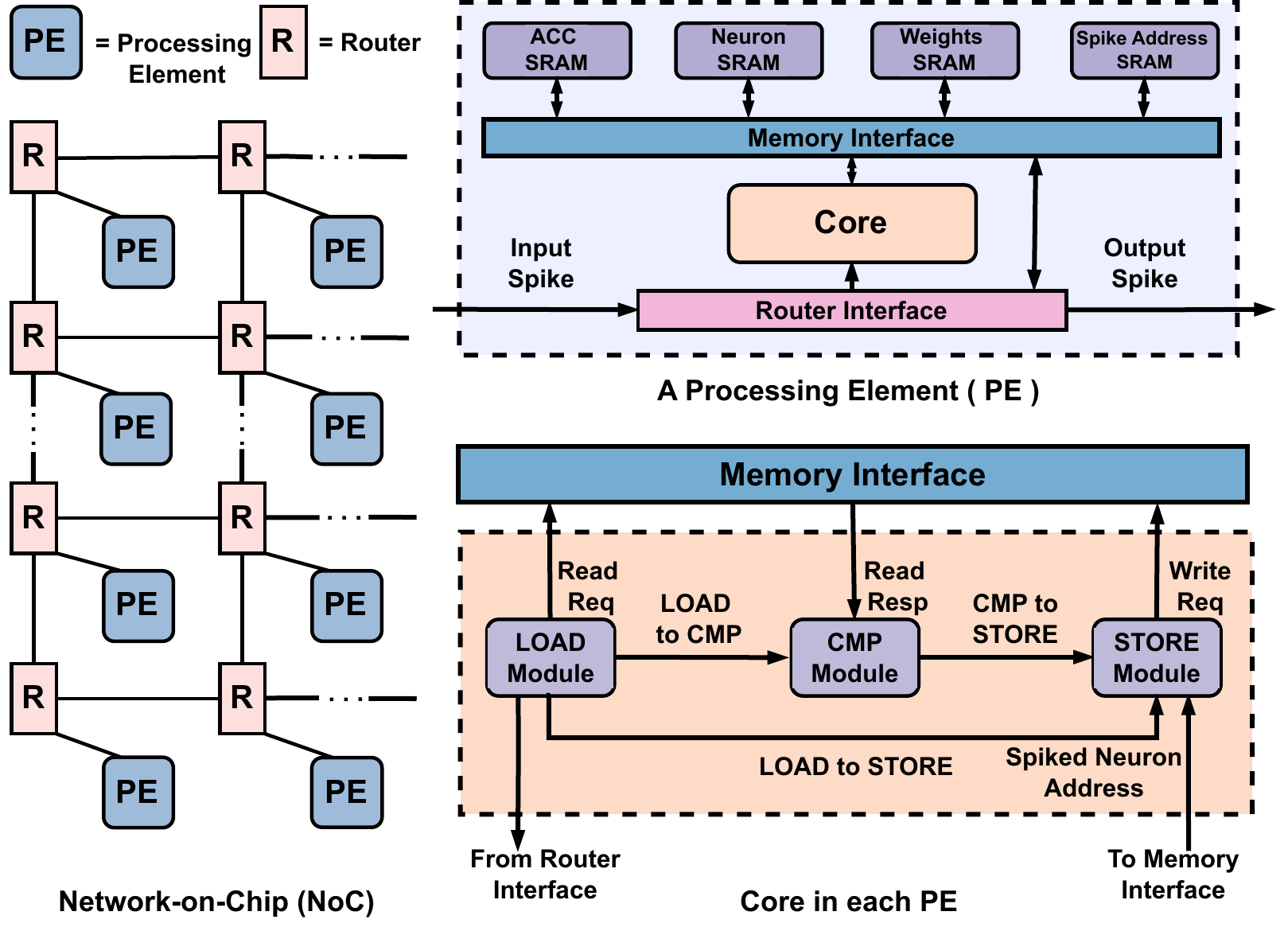}
    \caption{Overview of YOSO, a TTFS-based neuromorphic hardware \cite{yu2023ttfs}.}
    \label{fig:yoso}
\end{figure}

We evaluated the practical deployability of TNAS-ER by deploying it on YOSO \cite{yu2023ttfs,chu2020you}, a general-purpose neuromorphic hardware platform for TTFS SNN. This experiment aims to validate whether the searched architecture can be effectively executed on neuromorphic hardware under realistic system constraints. As shown in Figure \ref{fig:yoso}, YOSO consists of multiple Processing Elements (PEs) interconnected through a Network-on-Chip (NoC). Each PE integrates a compute core capable of parallel load, write, and compute operations, with clock-gated SRAMs for storing weights, neuron states, accumulated values, and spike information. The system operates at 54 MHz.

Before deployment, the trained TNAS-ER model is quantized, achieving 76.6\% WAR and 77.1\% UAR.
During hardware mapping, different layers are assigned to distinct PE groups to maximize parallelism. Each output feature map is partitioned into multiple regions according to the number of neurons, with each region processed by a single PE. This mapping strategy minimizes inter-PE data transactions, thereby reducing overall energy consumption. More details are provided in the Supplementary. Following the evaluation protocol of EMO~\cite{wu2020emo}, we measure only the resource usage during the emotion recognition phase, as the overhead of eye-region video acquisition lies outside network computation. 

\begin{figure}
    \centering
    \includegraphics[width=1.0\linewidth]{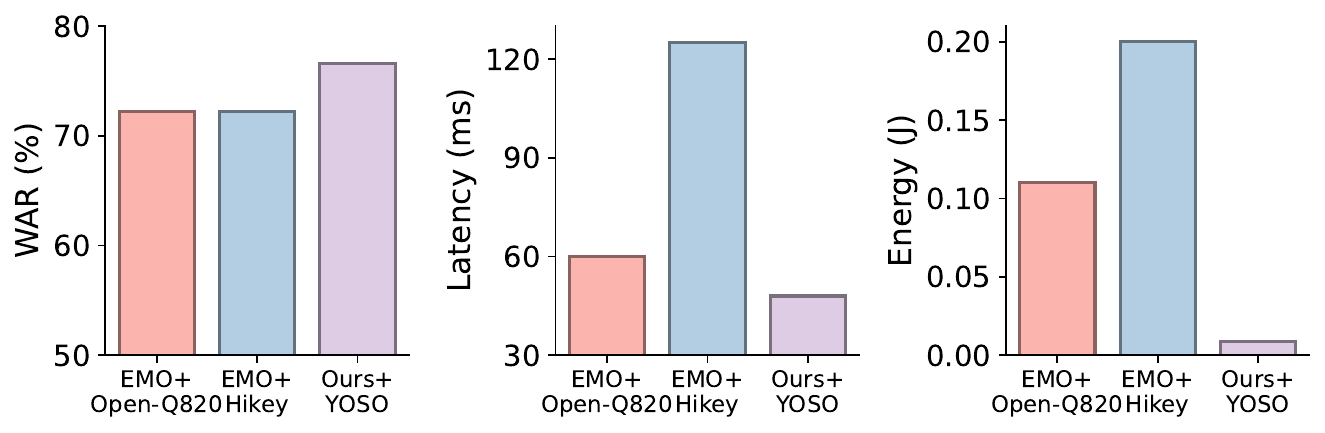}
    \caption{Hardware performance of different eye-based emotion recognition systems.}
    \label{fig:hardware}
\end{figure}

 The TNAS-ER + YOSO system achieves an energy consumption of 9.0 mJ and a latency of 48.0~ms. Since the latency is much shorter than the 0.4~s duration of event input, the system satisfies real-time processing requirements. Due to the lack of hardware evaluations in eye-based emotion recognition models, and none of them are TTFS-based, direct hardware performance comparisons are not feasible. The EMO system~\cite{wu2020emo}, which has been deployed on a wearable headset, is the only prior eye emotion recognition work that reports hardware performance. As shown in Figure~\ref{fig:hardware}, our system exhibits higher accuracy, lower energy consumption and shorter latency than the EMO system~\cite{wu2020emo}. These results indicate that the proposed method holds strong potential for practical deployment in future wearable devices and provide a reference baseline for future research on event-based eye emotion recognition.

\section{Conclusion}
In this paper, we present TNAS-ER, a NAS framework tailored to TTFS SNNs for efficient eye-based emotion recognition. TNAS-ER introduces an ANN-assisted search strategy that couples the TTFS SNN with its ReLU ANN counterpart to enable stable training and effective architecture exploration. Extensive experimental results demonstrate that TNAS-ER achieves state-of-the-art recognition performance with 1.6$\times$ fewer parameters, 5.6$\times$ fewer operations and 5.8 $\times$ shorter inference latency compared to existing ANN and SNN baselines. We further validate that TNAS-ER outperforms both manually designed and randomly searched architectures.
We also deploy TNAS-ER on the YOSO neuromorphic hardware and show the strong potential of the system for practical deployment. Overall, this work underscores the potential of architecture-level optimization in advancing spiking neural computing and paves the way for next-generation wearable and edge-intelligent systems.
{
    \small
    \bibliographystyle{ieeenat_fullname}
    \bibliography{main}
}

\end{document}